\documentclass[conference]{IEEEtran}
\ifCLASSINFOpdf
\else
\fi
\usepackage{listings}
\usepackage{stfloats}
\usepackage{color}
\usepackage{graphicx}
\usepackage{comment}
\usepackage{lipsum}
\usepackage{subfig}
\usepackage{multirow}
\usepackage{xspace}
\usepackage{amsmath}
\usepackage{bm}
\usepackage{tabularx}
\usepackage{hyperref}
\usepackage[section]{placeins}
\usepackage[utf8]{inputenc}
\usepackage{footnote}
\makesavenoteenv{tabular}
\makesavenoteenv{table}
\usepackage{dingbat}
\usepackage{wasysym}

\begin{document}

\newcommand{\todoinline}[1]{\todo[inline]{#1}}
\newcommand{\rarrow}{\ensuremath{\rightarrow}\xspace}
\newcommand{\blind}[1]{\texttt{\lipsum[#1]}}
\newcommand{\QuestionAnswering}{Question Answering\xspace}
\newcommand{\ResourceDescriptionFramework}{Resource Description Framework\xspace}
\newcommand{\NaturalLanguageProcessing}{Natural Language Processing\xspace}
\newcommand{\MachineTranslation}{Machine Translation\xspace}
\newcommand{\QA}{QA\xspace}
\newcommand{\Classifier}[1]{\boldsymbol{$M_#1$}\xspace}
\newcommand{\API}{API\xspace}
\newcommand{\GERBIL}{GERBIL\xspace}
\newcommand{\NLP}{NLP\xspace}
\newcommand{\Python}{Python\xspace}
\newcommand{\ML}{ML\xspace}
\newcommand{\MT}{MT\xspace}
\newcommand{\MLIR}{MLIR\xspace}
\newcommand{\CLEF}{CLEF\xspace}
\newcommand{\RDF}{RDF\xspace}
\newcommand{\dbo}[1]{\href{http://dbpedia.org/ontology/#1}{\texttt{dbo:#1}}}
\newcommand{\VQA}{VQA\xspace}
\newcommand{\SQA}{SQA\xspace}
\newcommand{\MC}{MC\xspace}
\newcommand{\HCI}{HCI\xspace}
\newcommand{\MRC}{MRC\xspace}
\newcommand{\ASR}{ASR\xspace}
\newcommand{\KGQA}{KGQA\xspace}
\newcommand{\ODQA}{ODQA\xspace}
\newcommand{\SPO}{SPO\xspace}
\newcommand{\TFIDF}{TFIDF\xspace}
\newcommand{\SPARQL}{SPARQL\xspace}
\newcommand{\RuBQ}{RuBQ\xspace}
\newcommand{\DeepPavlov}{DeepPavlov\xspace}
\newcommand{\QAnswer}{QAnswer\xspace}
\newcommand{\English}{English\xspace}
\newcommand{\German}{German\xspace}
\newcommand{\Wikidata}{Wikidata\xspace}
\newcommand{\Russian}{Russian\xspace}
\newcommand{\KnowledgeGraph}{KG\xspace}
\newcommand{\KG}{\KnowledgeGraph}
\newcommand{\XKnowSearch}{XKnowSearch!\xspace}
\newcommand{\BreXearch}{BreXearch\xspace}
\newcommand{\Table}{Table\xspace}
\newcommand{\Figure}{Figure\xspace}
\newcommand{\RU}{ru\xspace}
\newcommand{\EN}{en\xspace}
\newcommand{\refsec}[1]{Section~\ref{#1}\xspace}
\newcommand{\reffig}[1]{\Figure~\ref{#1}\xspace}
\newcommand{\reftab}[1]{\Table~\ref{#1}\xspace}
\newcommand{\QALLME}{QALL-ME\xspace}
\newcommand{\UTQA}{UTQA\xspace}
\newcommand{\BLEU}{BLEU\xspace}
\newcommand{\NIST}{NIST\xspace}
\newcommand{\CWQ}{CWQ\xspace}
\newcommand{\CFQ}{CFQ\xspace}
\newcommand{\Platypus}{Platypus\xspace} %\footnote{\url{(https://askplatyp.us}}
\newcommand{\RQ}[1]{RQ\ensuremath{_{#1}}}

\newcommand{\ie}{i.e.,~}
\newcommand{\incl}{incl.~}
\newcommand{\wrt}{w.r.t.~}
\newcommand{\st}{s.t.,~}
\newcommand{\eg}{e.g.,~}
\newcommand{\egPlain}{e.g.,}
\newcommand{\Eg}{For example,~}
\newcommand{\cf}{cf.,~}
\newcommand{\etc}{etc.\xspace}
\newcommand{\qq}[1]{``#1''}
\newcommand{\LRL}{LRL\xspace}
\newcommand{\LRLs}{LRLs\xspace}
\newcommand{\QALD}{QALD\xspace}
\newcommand{\URI}{URI\xspace}
\newcommand{\mKGQA}{mKGQA\xspace}
\newcommand{\LCQuAD}{LC-QuAD\xspace}
\newcommand{\WebQuestions}{WebQuestions\xspace}
\newcommand{\SimpleQuestions}{SimpleQuestions\xspace}
\newcommand{\CogComp}{TREC\xspace}
\newcommand{\TagMe}{TAGME\xspace}
\newcommand{\DBpedia}{DBpedia\xspace}
\newcommand{\Fscore}{F1 score\xspace}
\newcommand{\Precision}{Precision\xspace}
\newcommand{\Recall}{Recall\xspace}
\newcommand{\NDCG}{NDCG\xspace}
\newcommand{\WER}{WER\xspace}
\newcommand{\RTF}{RTF\xspace}
\newcommand{\QALDplus}{QALD-9-plus\xspace}
\newcommand{\BLI}{BLI\xspace}

\title{QALD-9-plus: A Multilingual Dataset for Question Answering over \DBpedia and \Wikidata\\Translated by Native Speakers}

\newcommand{\authorbox}[1]{
\begin{minipage}{0.48635\linewidth}\centering\vspace{0.25ex}
#1
\end{minipage}
}

\author{
\IEEEauthorblockN{Aleksandr Perevalov}
\IEEEauthorblockA{\authorbox{
Department of Computer Science and Languages\\
Anhalt University of Applied Sciences\\
Köthen (Anhalt), Germany\\
Email: aleksandr.perevalov@hs-anhalt.de}}
\and
\IEEEauthorblockN{Dennis Diefenbach}
\IEEEauthorblockA{\authorbox{
Laboratoire Hubert Curien\\
CNRS UMR 5516\\
Université de Lyon, France\\
Email: dennis.diefenbach@univ-st-etienne.fr}}
\and
\IEEEauthorblockN{Ricardo Usbeck}
\IEEEauthorblockA{\authorbox{
Department of Informatics\\
University of Hamburg\\
Hamburg, Germany\\
Email: ricardo.usbeck@uni-hamburg.de}}
\and
\IEEEauthorblockN{Andreas Both}
\IEEEauthorblockA{\authorbox{
Department of Computer Science and Languages\\
Anhalt University of Applied Sciences\\
Köthen (Anhalt), Germany\\
Email: andreas.both@hs-anhalt.de}}
}


%Telephone: (800) 555--1212, Fax: (888) 555--1212}


\maketitle

\begin{abstract}
The ability to have the same experience for different user groups (i.e., accessibility) is one of the most important characteristics of Web-based systems.
The same is true for Knowledge Graph Question Answering (KGQA) systems that provide the access to Semantic Web data via natural language interface.
While following our research agenda on the multilingual aspect of accessibility of KGQA systems, we identified several ongoing challenges.
One of them is the lack of multilingual KGQA benchmarks.
In this work, we extend one of the most popular KGQA benchmarks -- QALD-9 by introducing high-quality questions' translations to 8 languages provided by native speakers, and transferring the SPARQL queries of QALD-9 from DBpedia to Wikidata, s.t., the usability and relevance of the dataset is strongly increased.
Five of the languages -- Armenian, Ukrainian, Lithuanian, Bashkir and Belarusian -- to our best knowledge were never considered in KGQA research community before.
The latter two of the languages are considered as \qq{endangered} by UNESCO.
We call the extended dataset QALD-9-plus and made it available online\footnote{Figshare: \url{https://doi.org/10.6084/m9.figshare.16864273}. GitHub: \url{https://github.com/Perevalov/qald\_9\_plus}}.
\end{abstract}


% \fi
\IEEEpeerreviewmaketitle

\section{Introduction}

The core task of a Knowledge Graph Question Answering system is to represent a natural language question in the form of a structured query (\eg \SPARQL) to a knowledge graph (\KG).
In other words, \KGQA systems provide access to the data in KGs via a natural-language user interface, \st end users are not required to learn a particular query language for fetching data manually.
Obviously, the relevance (or accuracy) of the answers given by such system should strive to human performance and reduce labor costs for learning a particular query language; otherwise, the system is useless.
Many researchers are aiming at measuring and increasing the Question Answering (\QA) quality or the quality of a particular \KGQA sub-tasks, such as named entity linking (\eg \cite{DBLP:conf/icwe/DiefenbachSBC0A17}), expected answer type prediction (\eg \cite{smartPerevalov}), \etc
However, \emph{the accessibility\footnote{The accessibility for the Web is defined by W3C: \url{https://www.w3.org/standards/webdesign/accessibility}} characteristic of the \KGQA systems often stays overlooked}.
In this context, the perfect accessibility denotes an equivalent experience to all user groups of a particular \KGQA system.
Hence, such research questions as: \qq{How many people can really take advantage of the high-quality \KGQA system?} and \qq{Who are these people?} as well as \qq{How diverse they are?} are often left unnoticeable.

In this work, \emph{we focus on the multilingual aspect of accessibility}. 
The majority of research activities are dedicated to using the English language, however, there are approximately 7000 languages spoken in the world.
Moreover, many languages spoken by hundreds of thousands or even millions of people (\eg Udmurt, Bashkir, Belarusian) are not considered in the \KGQA research to this moment.
In the context of the multilingual aspect, increased accessibility provides the opportunity to use a \KGQA system effectively for the speakers of different languages (including low-resource and endangered ones\footnote{\cf \url{http://www.unesco.org/languages-atlas/en/atlasmap.html}}).
While doing preliminary research, we identified that multilingual \KGQA (\mKGQA) has several ongoing challenges. 
One of them is the lack of available benchmarks.
To the best of our knowledge, there are only 3 \mKGQA datasets in the research community (\QALD-9 \cite{QALD}, \RuBQ 2.0 \cite{rubq20}, and \CWQ \cite{cui2021multilingual}) and all of them are not fulfilling completely the practical needs of researchers and developers (see Section~\ref{sec:related_work}).
Hence, \emph{even if one develops an \mKGQA system, there are not so many opportunities for full-fledged evaluation}. 
\begin{comment}
\begin{figure*}[t!]
    \centering
    \includegraphics[width=\textwidth]{img/translation-illustration.pdf}
    \caption{Illustration of the translation process. TODO: redraw}
    \label{fig:translation-illustration}
    % source: https://docs.google.com/drawings/d/1eJL-xFtLUrbkQt6uKPvpZbKvUnbW_xxPoyACq-5m_mQ/edit?usp=sharing
\end{figure*}
\end{comment}
In this regard, it was decided to extend the well-known \QALD-9 dataset in order to enlarge its language coverage and improve the quality of non-English questions.
%The work was triggered by observing low-quality multilingual natural language representations of the questions that were identified previously.
The English questions from the dataset are considered as a reference.
To create the extension, a number of crowd workers were asked to translate the questions into their mother tongue manually (and validate translations by others).
As a result, 17 volunteer participants and 290 crowd workers from Amazon Mechanical Turk\footnote{\url{https://www.mturk.com/}} and Yandex Toloka\footnote{\url{https://toloka.yandex.com/}} translated questions into 8 languages, such as: German, Russian, French, Armenian, Belarusian, Lithuanian, Bashkir, and Ukrainian, the latter five of which, to our knowledge, have never been considered in the \KGQA context.
In addition, the \SPARQL queries in the dataset were transformed from \DBpedia to \Wikidata in order to increase the usability of the dataset.
We name the extended dataset as \qq{\QALDplus} and plan to continuously extend it over time.
Thus, \emph{the main contribution of this work} is that we introduce the \mKGQA benchmark with high-quality translations that covers 9 languages in total (including English) some of that were never covered in the \KGQA research community before. 
Moreover, Belarusian and Bashkir languages are considered as endangered by UNSECO\footnote{\cf \url{http://www.unesco.org/languages-atlas/}}.
We value our contribution as a small but very important step towards improving multilingual accessibility of \KGQA systems.
%Prior research has demonstrated that users of closely-related IR engines prefer to use their native language to search the information \cite{}.

This paper is structured as follows: in Section \ref{sec:related_work} we describe currently available multilingual \KGQA datasets, Section \ref{sec:qald_plus_creation} describes in details the process of creation and analysis of \QALDplus.
We demonstrate evaluation results of multilingual \KGQA systems on \QALDplus in Section \ref{sec:evaluation} and conclude the paper in Section \ref{sec:conclusion}.
%%%%%%%%%%%%%%%%%%%%%%%%%%%%%%%%%%%%%%%%%%%%%%%%%%%%%%%%%%%%%%%%%%%%%%%%%%%%%
%%%%%%%%%%%%%%%%%%%%%%%%%%%%%%%%%%%%%%%%%%%%%%%%%%%%%%%%%%%%%%%%%%%%%%%%%%%%%
\section{Multilingual \KGQA Benchmarks}\label{sec:related_work}
%%%%%%%%%%%%%%%%%%%%%%%%%%%%%%%%%%%%%%%%%%%%%%%%%%%%%%%%%%%%%%%%%%%%%%%%%%%%%

Today, the research in the field of \KGQA is strongly dependent on data, and it suffers from the lack of multilingual benchmarks \cite{loginovaMT,LackOfQADatasetsBioMed}. %more citations 
To the best of our knowledge, only three \KGQA benchmarks exist that tackle multiple languages: \QALD \cite{QALD}, \RuBQ \cite{rubq20}, and \CWQ \cite{cui2021multilingual}.

\QALD-9 contains 558 questions incorporating information of the DBpedia knowledge base\footnote{\url{https://www.dbpedia.org/}} \cite{dbpedia} where for each question the following is given: a textual representation in multiple languages, the corresponding \SPARQL query (over DBpedia), the answer entity \URI, and the answer type. 
The dataset has become a benchmark for many research studies in \QA (\eg \cite{SurveyOnChallengesOfQA,DBLP:conf/icwe/DiefenbachSBC0A17,sorokin2018modeling,QAnswer,MQALD}).

\RuBQ 2.0 is a \KGQA dataset over \Wikidata\footnote{\url{https://www.wikidata.org/}} \cite{Wikidata} that contains 2910 questions. 
The questions are represented in native \Russian language and machine-translated to \English language.
Additionally, it contains a list of entities, relations, answer entities, \SPARQL queries, answer-bearing paragraphs, and query type tags. 

\CWQ is a recently published \KGQA dataset over \Wikidata that is based on \CFQ data \cite{keysers2019measuring}.
\CWQ contains questions in Hebrew, Kannada, Chinese, and English languages.
All the non-English questions were obtained using machine translation with several rule-based adjustments.
It has a well-detailed structure including: question with highlighted entities, original \SPARQL query over Freebase \cite{bollacker2008freebase} (from \CFQ), \SPARQL query over \Wikidata (introduced in \CWQ), textual representations of a question in four aforementioned languages, and additional fields.

Despite the considered benchmarks contain questions in multiple languages, these multilingual representations were either machine-translated (\RuBQ 2.0, \CWQ) or have doubtful quality (\QALD-9, see Section \ref{ssec:qualitative_analysis_qald9}).
In contrast, the dataset that we present is covering 9 languages (including English) and was created by native speakers.
%%%%%%%%%%%%%%%%%%%%%%%%%%%%%%%%%%%%%%%%%%%%%%%%%%%%%%%%%%%%%%%%%%%%%%%%%%%%%%%%%%%%
%%%%%%%%%%%%%%%%%%%%%%%%%%%%%%%%%%%%%%%%%%%%%%%%%%%%%%%%%%%%%%%%%%%%%%%%%%%%%%%%%%%%
\section{\QALDplus Benchmark Creation Process}\label{sec:qald_plus_creation}
%%%%%%%%%%%%%%%%%%%%%%%%%%%%%%%%%%%%%%%%%%%%%%%%%%%%%%%%%%%%%%%%%%%%%%%%%%%%%%%%%%%%
\subsection{Qualitative Analysis of \QALD-9}\label{ssec:qualitative_analysis_qald9}
%%%%%%%%%%%%%%%%%%%%%%%%%%%%%%%%%%%%%%%%%%%%%%%%%%%%%%%%%%%%%%%%%%%%%%%%%%%%%%%%%%%%
While working with \QALD-9 the following flaws in the data were identified.

Firstly, there is poor translation quality of the majority of questions in languages other than English.
The authors, as native speakers of German and Russian, were capable to identify them for the respective languages.
The question: \qq{Which subsidiary of TUI Travel serves both Glasgow and Dublin?} has the following translation to Italian: \qq{Quale società sussidiaria di TUI Travel serve sia Dortmund che Dublino?}.
While not being a native speaker of Italian, it is obvious that two different cities are used in the original question (Glasgow) and in its translated version (Dortmund).
It is worth noting, that the corresponding \SPARQL query uses Glasgow, as in the original English question.
Additionally, an analysis was performed for the German questions by an English teacher from Germany, rating the quality of the translation from English to German. 
It scores only 4.09 of 5.00 (5.00 is the highest rating) where 29 question received a rating of 1, 40 a rating of 2 (87 of 3, 97 of 4, and 305 of 5). 
Hence, this qualitative rating of the questions' translation to German shows that the translation quality is not sufficient.

Secondly, the meaning of some questions is ambiguous, for example: \qq{How often did Jane Fonda marry?}, here, the actual meaning is probably how many times.
Another ambiguous example of a question is: \qq{Who is the heaviest player of the Chicago Bulls?}, in this case, the time span of the question is unclear -- the heaviest in the history or currently playing.
The corresponding \SPARQL query for the question \qq{Give me all films produced by Steven Spielberg with a budget of at least \$80 million.} contains the property \dbo{director}, however, the question asks for \dbo{producer}.

In our work, we focus on the first part of the dataset's flaws -- poor translation quality. 
The original \DBpedia queries were not changed, but were transformed to \Wikidata.
In the next sections, we describe the translation process and the process of migration of the \SPARQL queries from the \DBpedia KG to the \Wikidata KG.
%%%%%%%%%%%%%%%%%%%%%%%%%%%%%%%%%%%%%%%%%%%%%%%%%%%%%%%%%%%%%%%%%%%%%%%%%%%%%%%%%%%%
%%%%%%%%%%%%%%%%%%%%%%%%%%%%%%%%%%%%%%%%%%%%%%%%%%%%%%%%%%%%%%%%%%%%%%%%%%%%%%%%%%%%
\subsection{Translation of Questions}
%%%%%%%%%%%%%%%%%%%%%%%%%%%%%%%%%%%%%%%%%%%%%%%%%%%%%%%%%%%%%%%%%%%%%%%%%%%%%%%%%%%%

The translation process was executed in a crowdsourcing manner with the following settings: (1) the crowd workers had to translate the questions from English to their mother tongue, (2) each crowd worker was provided with a prepared subset of questions selected from different parts of the dataset to avoid possible biases (one crowd worker translated at least 10 questions), (3) the usage of machine translation tools was prohibited (only dictionaries were allowed), (4) the crowd workers were not aware of the existing multilingual representations from the \QALD-9 dataset, and (5) the crowd workers were not aware of the other's tasks. 
%The translation process is formalized in Figure~\ref{fig:translation-illustration}.
In total, 17 volunteers and 290 crowd workers from Amazon Mechanical Turk and Yandex Toloka speaking 8 languages took part in the translation process.
The coverage of translations was set to $\geq$ 2.

After the translations were obtained, the validation process was executed.
During the validation, the crowd workers were provided with the original question and two translation options.
A crowd worker had to select one of the following options: (1) no translations are correct, (2) first translation is correct, (3) second translation is correct, or (4) both translations are correct.

Despite the volunteers, the tasks to the crowd workers were assigned according to the main country of language's origin (\eg one has to be from France to translate from English to French).
Thus, the questions were translated by native speakers of these different languages and additionally multiple times validated.

\begin{figure*}[t]
\centering
\begin{minipage}[t]{0.44\linewidth}
\begin{lstlisting}[captionpos=b]
# DBpedia
SELECT DISTINCT ?date 
WHERE { 
  dbr:Finland dbp:accessioneudate ?date .
}
\end{lstlisting}
\end{minipage}
\begin{minipage}[t]{0.05\linewidth}
\ 
\end{minipage}
\begin{minipage}[t]{0.44\linewidth}
%\begin{minted}{sparql}
\begin{lstlisting}[captionpos=b]
# Wikidata
SELECT DISTINCT ?date 
WHERE { 
  wd:Q33 p:P463 ?membership . 
  ?membership pq:P580 ?date . 
  ?membership ps:P463 wd:Q458 . 
}
\end{lstlisting}
%\end{minted}
\end{minipage}
\caption{An example of different basic graph patterns in DBpedia and Wikidata for question: \qq{When did Finland join the EU?}. Prefixes are omitted (we used the typical RDF prefixes as shown at \href{https://prefix.cc/popular/all}{https://prefix.cc/popular/all}).}
\label{fig:SPARQL}
\end{figure*}

\begin{table*}[t]
\centering
\caption{The number of \qq{gold standard} answer sets computed over \DBpedia that have less or equal corresponding intersection rate between \QALD-9 and \QALDplus, \ie how many \qq{gold standard} answer sets have the intersection rate less than X\%?}
\label{tab:answer_set_similarity}
\begin{tabular}{|l|c|c|c|c|}
\hline
 & \textbf{$\leq$ 25\%} & \textbf{$\leq$ 50\%} & \textbf{$\leq$ 75\%} & \textbf{$\leq$ 100\%} \\
 \hline
\textbf{Train}         & 183                        & 227                        & 269                        & 408                         \\
\hline
\textbf{Test}          & 69                         & 86                         & 98                         & 150 \\ \hline

\end{tabular}
\end{table*}

\begin{table*}[t]
\centering
  \caption{Overview of the extended \QALDplus.}
  \label{tab:qald_plus_overview}
\begin{tabular}{|l|c|c|c|c|c|c|c|c|c|c|c|c|}
\hline
               & \textbf{en} & \textbf{de} & \textbf{fr} & \textbf{ru} & \textbf{uk} & \textbf{lt} & \textbf{be} & \textbf{ba} & \textbf{hy} & \textbf{DBpedia} & \textbf{Wikidata} & \textbf{\# questions} \\ \hline
\textbf{Train} & 408         & 543         & 260          & 1203        & 447         & 468         & 441         & 284         & 80          & 408              & 371               & 408                   \\ \hline
\textbf{Test}  & 150         & 176         & 26          & 348         & 176         & 186         & 155         & 117         & 20          & 150              & 136               & 150                   \\ \hline
\end{tabular}
\end{table*}

%%%%%%%%%%%%%%%%%%%%%%%%%%%%%%%%%%%%%%%%%%%%%%%%%%%%%%%%%%%%%%%%%%%%%%%%%%%
\subsection{Migration from \DBpedia to \Wikidata}
%%%%%%%%%%%%%%%%%%%%%%%%%%%%%%%%%%%%%%%%%%%%%%%%%%%%%%%%%%%%%%%%%%%%%%%%%%%

To extend the usability of the dataset, we ported the corresponding \QALD-9 queries from \DBpedia to the \Wikidata KG. 
This process was carried out and validated in-house by 3 experienced computer scientists.
Although semi-automatic scripts for retrieving entity and property mappings were partly used to speed up the process, it was very labor-intensive.
In particular, we found that a simple property path in \DBpedia is not necessarily simple considering the \Wikidata KG, which complicated the transformation process.
For example, the \DBpedia query for the question \qq{When did Finland join the EU?}, consists of one triple (\DBpedia) while three triples are required to fetch the same information from the \Wikidata KG (see Figure~\ref{fig:SPARQL}).

Several \DBpedia \SPARQL queries were not portable to \Wikidata due to the absence of the corresponding information (\eg \qq{How many calories does a baguette have?}).
In this regard, 51 queries could not be transferred to \Wikidata, since the information available in the \DBpedia KG was not available in the \Wikidata KG. 
% TODO: update Wikidata?
The \qq{gold standard} answers over \Wikidata for the questions were obtained by executing the corresponding \SPARQL queries. 
The updated \qq{gold standard} answers over \DBpedia were obtained the same way.

%TODO: how updated gold standard DBpedia has changed?

\begin{comment}
\begin{table*}[t]
    \caption{Number of \QALD-9 questions improved in the \QALDplus dataset}
    \label{tab:qald_plus_changes}
    \centering
    \begin{tabular}{c|c|c|c}
         &  \textbf{de} & \textbf{ru} & \textbf{fr} \\\hline
         \textbf{number of question in train and test in \QALD-9} & & & \\\hline
         \textbf{number of changed question in train and test in \QALDplus} & & & \\\hline
    \end{tabular}
\end{table*}
\end{comment}

We compared the answer-sets of the SPARQL queries of the old (\QALD-9) and the updated (\QALDplus) datasets (see Table \ref{tab:answer_set_similarity}). Given the table, nearly half of the \qq{gold standard} answer sets for both train and test splits has less than 50\% similarity, while comparing \QALD-9 and \QALDplus.
That is caused by changes in the \DBpedia that may occur due to several reasons.
For example, historical reasons, such as the change of president, governor \etc, play a role, but also the data model has changed over time, or missing information was updated or added.
This shows also the importance of the answer sets in the \KGQA datasets since they enable comparability between different versions of a knowledge base.
The up-to-date information should always be retrieved via the corresponding \SPARQL queries.

%%%%%%%%%%%%%%%%%%%%%%%%%%%%%%%%%%%%%%%%%%%%%%%%%%%%%%%%%%%%%%%%%%%
\subsection{Dataset Statistics}
%%%%%%%%%%%%%%%%%%%%%%%%%%%%%%%%%%%%%%%%%%%%%%%%%%%%%%%%%%%%%%%%%%%

%The overview of the dataset statistics is present in Table~\ref{tab:qald_plus_overview}.
%Given the numbers, it is obvious that some questions are covered more than once, \ie there is more than one translation for a particular question.
%For example, there are 1294 Russian translations available while only 408 unique questions exist in the training subset (\ie overlap -- 3.2 Russian translations per one question).
%This enables the dataset users to address also the paraphrasing task.

%We consider Armenian, Belarusian, Lithuanian, and Ukrainian as low-resource languages \wrt the ongoing research in \NLP and \QuestionAnswering.
%German, Russian, Italian, and French are considered to be well-researched but still far away from English language regarding the availability of datasets and tools.
%In addition, as the translations were done by native speakers, they can be used as a \qq{gold standard} for the evaluation of machine translation tools, too.

%Consequently, due to the translation by native speakers, several preexisting question translations of \QALD-9 were changed for German and Russian to improve the quality.
%Table \ref{tab:qald_plus_changes} shows these changes.

The overview of the dataset statistics is present in Table~\ref{tab:qald_plus_overview}.\\
We extended the dataset with 4,930 new question translations for different languages. 
In this regard, the majority of the multilingual representations of \QALD-9 were replaced by the new ones.
Note that for some languages like Russian we could collect multiple translations for some questions (\ie 2.9 Russian translations per one question in training set) while for other languages like Armenian, Bashkir, French we provide only a partial converge of the original QALD-9 questions. 
All translations were carried out by native speakers.\\
In terms of the SPARQL queries, we extended the dataset with 507 new queries over \Wikidata. 
The queries were manually created by the authors of the paper.

\begin{table*}[t]
\centering
\caption{Comparison of quantitative text features between \QALD-9 and \QALDplus (both train and test subsets were considered)}
\label{tab:qualitative_analysis}
\begin{tabular}{|l|c|c|c|c|c|c|c|c|}
\hline
\multirow{2}{*}{} & \multicolumn{2}{c|}{\textbf{English (not changed)}} & \multicolumn{2}{c|}{\textbf{German}} & \multicolumn{2}{c|}{\textbf{Russian}} & \multicolumn{2}{c|}{\textbf{French}} \\ \cline{2-9} & \multicolumn{1}{c|}{\textbf{QALD-9}} & \multicolumn{1}{c|}{\textbf{\QALDplus}} & \multicolumn{1}{c|}{\textbf{QALD-9}} & \multicolumn{1}{c|}{\textbf{\QALDplus}} & \multicolumn{1}{c|}{\textbf{QALD-9}} & \multicolumn{1}{c|}{\textbf{\QALDplus}} & \multicolumn{1}{c|}{\textbf{QALD-9}} & \multicolumn{1}{c|}{\textbf{\QALDplus}} \\ \hline
\textbf{Average syllables per word} & \multicolumn{2}{c|}{1.32}  & 1.86 & 1.87 & 2.13 & 2.31 & 1.51 & 1.55 \\ \hline
\textbf{Average word length} & \multicolumn{2}{c|}{4.77} & 5.64 & 5.68 & 5.66 & 6.12 & 5.07 & 5.18 \\ \hline
\textbf{Average sentence length} & \multicolumn{2}{c|}{33.92} & 35.65 & 39.14 & 19.80 & 36.66 & 36.35 & 39.44 \\ \hline
\textbf{Average words per sentence} & \multicolumn{2}{c|}{7.06} & 6.32 & 6.89 & 3.49 & 5.99 & 7.18 & 7.62\\ \hline
\textbf{Type Token Ratio} & \multicolumn{2}{c|}{0.30} & 0.33 & 0.33 & 0.32 & 0.49 & 0.32 & 0.49 \\ \hline
\end{tabular}
\end{table*}
\newcommand{\notsupported}{\multicolumn{2}{c|}{\textit{not supported}}}
\newcommand{\langcol}[1]{\multicolumn{2}{c|}{\textbf{#1}}}
\begin{table*}[t]
\centering
\caption{Evaluation results on \QALDplus with \GERBIL system. Only multilingual \KGQA systems were used. The first value of a cell corresponds to Micro \Fscore, the second one corresponds to Macro \Fscore \cite{Gerbil}.}
%\footnote{There is no full coverage of all questions for French language (see Table \ref{tab:qald_plus_overview})}
\label{tab:evaluation}
\begin{tabular}{|l|c|c|c|c|c|c|c|c|}
\hline
\multicolumn{9}{|c|}{\textbf{QAnswer}} \\ \hline 
    \multirow{2}{*}{} & \langcol{English} & \langcol{German} & \langcol{Russian} & \langcol{French} \\ \cline{2-9}
    & Micro F1 & Macro F1 & Micro F1 & Macro F1 & Micro F1 & Macro F1 & Micro F1 & Macro F1 \\\hline
\textbf{\Wikidata test}  & 0.1002 & 0.4459 & 0.0291 & 0.3171 & 0.0159 & 0.2143 & 0.1190 & 0.2300 \\ \hline
\textbf{\Wikidata train} & 0.0866 & 0.4009 & 0.0587 & 0.3009 & 0.0382 & 0.2264 & 0.0926 & 0.2777 \\ \hline
\textbf{\DBpedia test}  & 0.0527 & 0.3039 & 0.0219 & 0.1998 & 0.0153 & 0.0957 & 0.0406 & 0.1506 \\ \hline
\textbf{\DBpedia train} & 0.1016 & 0.3416 & 0.1102 & 0.1807 & 0.0029 & 0.0436 & 0.0969 & 0.1968 \\ \hline
\multicolumn{9}{|c|}{\textbf{DeepPavlov}} \\ \hline
\textbf{\Wikidata test}  & 0.0013 & 0.1240 & \notsupported & 0.0005 & 0.0870 & \notsupported \\ \hline
\textbf{\Wikidata train} & 0.0017 & 0.1032 & \notsupported & 0.0009 & 0.0993 & \notsupported \\ \hline
\multicolumn{9}{|c|}{\textbf{Platypus}} \\ \hline
\textbf{\Wikidata test}  & 0.0126 & 0.1503 & \notsupported & \notsupported & 0.0000 & 0.0417 \\ \hline
\textbf{\Wikidata train} & 0.0028 & 0.1070 & \notsupported & \notsupported & 0.0063 & 0.0750 \\           \hline           
\end{tabular}
\end{table*}

\begin{table*}[t]
\centering
\caption{Evaluation results on \QALDplus with original \QALD-9 multilingual translations. The table structure is similar to Table \ref{tab:evaluation}. Only overlapping languages (German, Russian, French) are used for the better comparability (English was not changed).}
\label{tab:evaluation_original}
\begin{tabular}{|l|c|c|c|c|c|c|}
\hline
\multicolumn{7}{|c|}{\textbf{QAnswer}} \\ \hline 
    \multirow{2}{*}{} & \langcol{German} & \langcol{Russian} & \langcol{French} \\ \cline{2-7}
    & Micro F1 & Macro F1 & Micro F1 & Macro F1 & Micro F1 & Macro F1 \\\hline
\textbf{\Wikidata test}  & 0.0291 & 0.3171 & 0.0157 & 0.2059 & 0.1190 & 0.2300 \\ \hline
\textbf{\Wikidata train} & 0.0587 & 0.3013 & 0.0383 & 0.2264 & 0.1465 & 0.2813 \\ \hline
\textbf{\DBpedia test}  & 0.0219 & 0.1998 & 0.0154 & 0.1023 & 0.0406 & 0.1506 \\ \hline
\textbf{\DBpedia train} & 0.1087 & 0.1807 & 0.0029 & 0.0461 & 0.0190 & 0.1680 \\ \hline
\multicolumn{7}{|c|}{\textbf{DeepPavlov}} \\ \hline
\textbf{\Wikidata test}  & \notsupported & 0.0005 & 0.0870 & \notsupported \\ \hline
\textbf{\Wikidata train} & \notsupported & 0.0009 & 0.0993 & \notsupported \\ \hline
\multicolumn{7}{|c|}{\textbf{Platypus}} \\ \hline
\textbf{\Wikidata test}  &  \notsupported & \notsupported & 0.0000 & 0.0417 \\ \hline
\textbf{\Wikidata train} &  \notsupported & \notsupported & 0.0015 & 0.1004 \\           \hline           
\end{tabular}
\end{table*}

%%%%%%%%%%%%%%%%%%%%%%%%%%%%%%%%%%%%%%%%%%%%%%%%%%%%%%%%%%%%%%%%%%%
\subsection{Quantitative and Qualitative Question Analysis}\label{ssec:quantitative_analysis}
%%%%%%%%%%%%%%%%%%%%%%%%%%%%%%%%%%%%%%%%%%%%%%%%%%%%%%%%%%%%%%%%%%%

We used several language-agnostic quantitative text analysis measures in order to observe the change between \QALD-9 and \QALDplus textual representations \wrt the descriptive statistics.
Consequently, we analyzed only those languages that were present in both \QALD-9 and \QALDplus.
To do this, the LinguaF library\footnote{\url{https://github.com/Perevalov/LinguaF}} for Python was used.
The results of comparison are demonstrated in Table \ref{tab:qualitative_analysis}.

Given the data, it is obvious, that for all the \QALDplus translations longer words were used.
In addition, there are more words in the \QALDplus translated questions and consequently, the translated question length was increased.
The Type-Token Ratio (TTR) that corresponds to the ratio between unique words and total number of words was also increased for Russian and French, and not changed for German.
This analysis shows us that the \QALDplus multilingual representations are more complex and rich in comparison to the original \QALD-9 translations.
As the translations were done by native speakers, we consider this results as an implicit marker of improved translation quality.
Additionally, a well-experienced English teacher from Germany did an evaluation of the translation quality. 
The original dataset was rated with 4.09 points out of 5. 
The new German translations of the questions were also evaluated, and only 19 question translations were optimized in the final validation process step (where 4 changes addressed typos and missing quotation marks).
Hence, we can also conclude that the new German translations have a very high quality (5 out of 5 according to the previous rating). 

%%%%%%%%%%%%%%%%%%%%%%%%%%%%%%%%%%%%%%%%%%%%%%%%%%%%%%%%%%%%%%%%%%%%%%%%%%%%%%%%%%%%%%%%%%
\subsection{Impact and Usability}
%%%%%%%%%%%%%%%%%%%%%%%%%%%%%%%%%%%%%%%%%%%%%%%%%%%%%%%%%%%%%%%%%%%%%%%%%%%%%%%%%%%%%%%%%%
It was decided to keep the \QALD-JSON format\footnote{\url{https://github.com/dice-group/gerbil/wiki/Question-Answering\#web-service-interface}} for the \QALDplus as it enables the researchers to reuse their systems for the evaluation (\eg \GERBIL \cite{Gerbil}). 
Hence, researchers and developers are enabled to benchmark and compare their \KGQA systems in both mono- and multilingual settings over \DBpedia and \Wikidata.
As there are several alternative translations provided, it is possible to develop techniques for paraphrasing and evaluating machine translation quality.
The same is true for the \SPARQL queries over \DBpedia and \Wikidata.
\section{Evaluation of Multilingual KGQA Systems}\label{sec:evaluation}

We evaluated our \QALDplus dataset on currently available multilingual \KGQA systems: \QAnswer\cite{QAnswer}, \DeepPavlov\cite{deeppavlov} and \Platypus\cite{tanon2018demoing}.
\QAnswer supports 11 languages (en, de, fr, it, es, pt, nl, zh, ar, ja, ru) and works over \DBpedia and \Wikidata, \DeepPavlov\ -- 2 languages (en, ru) and works over \Wikidata, and \Platypus\ -- 2 languages (en, fr) and works over \Wikidata.
Consequently, we decided to evaluate the systems on the languages that are present in \QALDplus and are supported by them natively.
The evaluation was done using the \GERBIL system, the URIs of the experimental runs with detailed data (see the online appendix) are stored in the README.md file of the dataset.
The results are demonstrated in Table~\ref{tab:evaluation}.

From the results we see that \QAnswer has strong dominance \wrt English and French.
\DeepPavlov performed better on the \Wikidata test subset in Russian, however, \QAnswer is better on the corresponding train subset.
The results for German language and \DBpedia subset are not comparable due to the lack of its support.

Additionally, the original multilingual questions from \QALD-9 were used in the evaluation while having updated \SPARQL queries from \QALDplus (see Table \ref{tab:evaluation_original}).
For simpler comparability, only languages that were used for the \QALDplus were selected (\ie German, Russian, and French).
Given the results, no particular tendencies \wrt the quality difference between \QALD-9 and \QALDplus were identified.
Hence, while having a significantly different quantitative text measures between the original and extended translations (see Section \ref{ssec:quantitative_analysis}), the actual \QA quality was not changed significantly.
We hypothesize, that such result may be caused by the high robustness of the considered \KGQA systems to the grammatical correctness of the questions or the still very keyword-dependent behavior of the considered \KGQA systems.
Together with the robustness, the systems may also be capable of answering only a particular subset of questions, that might be the reason why the quality values for some experiments did not change.
However, this aspect has to be investigated further and is beyond the scope of this paper.

%%%%%%%%%%%%%%%%%%%%%%%%%%%%%%%%%%%%%%%%%%%%%%%%%%%%%%%%%%%%%%%%%%%%%%%%%%%%%%%%%%%
%%%%%%%%%%%%%%%%%%%%%%%%%%%%%%%%%%%%%%%%%%%%%%%%%%%%%%%%%%%%%%%%%%%%%%%%%%%%%%%%%%%
\section{Conclusion and Future Work}\label{sec:conclusion}
%%%%%%%%%%%%%%%%%%%%%%%%%%%%%%%%%%%%%%%%%%%%%%%%%%%%%%%%%%%%%%%%%%%%%%%%%%%%%%%%%%%

In this paper, we presented a new benchmark for KGQA called \QALDplus.
It is based on \QALD-9 and was extended by creating translations from its English questions to German, French, Russian, Armenian, Belarusian, Lithuanian, Bashkir, and Ukrainian.
The translations were done by native speakers of corresponding languages in a crowdsourcing setting.
In addition, the \DBpedia \SPARQL queries from \QALD-9 were transferred to \Wikidata to improve the usability of the data.
As \QALDplus contains multiple text representations for several languages and the questions are multilingual (\ie parallel corpus), it enables researchers to address the paraphrasing and machine translation tasks.
In addition, paraphrasing may be done on the \SPARQL query level (\ie from DBpedia to Wikidata).
\QALDplus is keeping the \QALD-JSON format in order to be reusable by the research community.
The quantitative and qualitative comparison of the \QALD-9 and \QALDplus multilingual question representations illustrated that the quality of the translation was improved.
We consider \QALDplus as a significant contribution to the multilingual KGQA research community that creates wider possibilities for evaluation of \KGQA systems.

By evaluation on three state-of-the-art KGQA systems using \QALDplus, we demonstrated that the extension to other KGs, namely \Wikidata, and improving the translations gave new insights into these systems' performance.

In the future, we will increase the coverage of languages (\eg Bashkir, Armenian, and French) and extend the number of languages in the dataset.
We also will enlarge the number of questions in the dataset, extend their meta-information (\eg expected answer type, named entities \etc), and align \SPARQL queries with the newest instance of \DBpedia.

\section*{Acknowledgment}

Authors of the paper would like to thank all the contributors involved in the translation of the dataset, specifically: Konstantin Smirnov, Mikhail Orzhenovskii, Andrey Ogurtsov, Narek Maloyan, Artem Erokhin, Mykhailo Nedodai, Aliaksei Yeuzrezau, Anton Zabolotsky, Artur Peshkov, Vitaliy Lyalin, Artem Lialikov, Gleb Skiba, Vladyslava Dordii, Polina Fominykh, Tim Schrader, Susanne Both, and Anna Schrader.
Additionally, authors would like to give thanks to Open Data Science community\footnote{\url{https://ods.ai/}} for connecting data science enthusiasts all over the world.




\bibliographystyle{IEEEtran}
\bibliography{references}

\begin{thebibliography}{10}
\providecommand{\url}[1]{#1}
\csname url@samestyle\endcsname
\providecommand{\newblock}{\relax}
\providecommand{\bibinfo}[2]{#2}
\providecommand{\BIBentrySTDinterwordspacing}{\spaceskip=0pt\relax}
\providecommand{\BIBentryALTinterwordstretchfactor}{4}
\providecommand{\BIBentryALTinterwordspacing}{\spaceskip=\fontdimen2\font plus
\BIBentryALTinterwordstretchfactor\fontdimen3\font minus
  \fontdimen4\font\relax}
\providecommand{\BIBforeignlanguage}[2]{{%
\expandafter\ifx\csname l@#1\endcsname\relax
\typeout{** WARNING: IEEEtran.bst: No hyphenation pattern has been}%
\typeout{** loaded for the language `#1'. Using the pattern for}%
\typeout{** the default language instead.}%
\else
\language=\csname l@#1\endcsname
\fi
#2}}
\providecommand{\BIBdecl}{\relax}
\BIBdecl

\bibitem{DBLP:conf/icwe/DiefenbachSBC0A17}
D.~Diefenbach, K.~Singh, A.~Both, D.~Cherix, C.~Lange, and S.~Auer, ``The
  {Qanary} ecosystem: Getting new insights by composing question answering
  pipelines,'' in \emph{Web Engineering - 17th International Conference, {ICWE}
  2017, Rome, Italy, June 5-8, 2017, Proceedings}, ser. Lecture Notes in
  Computer Science, J.~Cabot, R.~D. Virgilio, and R.~Torlone, Eds., vol.
  10360.\hskip 1em plus 0.5em minus 0.4em\relax Springer, 2017, pp. 171--189.

\bibitem{smartPerevalov}
\BIBentryALTinterwordspacing
A.~Perevalov and A.~Both, ``Augmentation-based answer type classification of
  the {SMART} dataset,'' in \emph{Proceedings of the SeMantic AnsweR Type
  prediction task {(SMART)} at {ISWC} 2020}, ser. {CEUR} Workshop Proceedings,
  vol. 2774.\hskip 1em plus 0.5em minus 0.4em\relax CEUR-WS.org, 2020, pp.
  1--9. [Online]. Available: \url{http://ceur-ws.org/Vol-2774/paper-01.pdf}
\BIBentrySTDinterwordspacing

\bibitem{QALD}
R.~Usbeck, R.~H. Gusmita, A.~N. Ngomo, and M.~Saleem, ``9th challenge on
  question answering over linked data {(QALD-9)},'' 2018.

\bibitem{rubq20}
I.~Rybin, V.~Korablinov, P.~Efimov, and P.~Braslavski, ``{RuBQ} 2.0: An
  innovated russian question answering dataset,'' in \emph{The Semantic
  Web}.\hskip 1em plus 0.5em minus 0.4em\relax Cham: Springer International
  Publishing, 2021, pp. 532--547.

\bibitem{cui2021multilingual}
R.~Cui, R.~Aralikatte, H.~Lent, and D.~Hershcovich, ``Multilingual
  compositional {Wikidata} questions,'' \emph{arXiv:2108.03509}, 2021.

\bibitem{loginovaMT}
E.~Loginova, S.~Varanasi, and G.~Neumann, ``Towards end-to-end multilingual
  question answering,'' \emph{Information Systems Frontiers (ISF)}, vol.~22,
  pp. 1--14, 3 2020.

\bibitem{LackOfQADatasetsBioMed}
A.~Neves, A.~Lamurias, and F.~Couto, ``Biomedical question answering using
  extreme multi-label classification and ontologies in the multilingual
  panorama,'' in \emph{Semantic Indexing and Information Retrieval for Health
  Held in conjunction with the 42nd European Conference on Information
  Retrieval (SIIRH@ECIR)}, 2020.

\bibitem{dbpedia}
S.~Auer, C.~Bizer, G.~Kobilarov, J.~Lehmann, R.~Cyganiak, and Z.~Ives,
  ``{DBpedia}: A nucleus for a web of open data,'' in \emph{The semantic
  web}.\hskip 1em plus 0.5em minus 0.4em\relax Springer, 2007.

\bibitem{SurveyOnChallengesOfQA}
K.~Höffner, S.~Walter, E.~Marx, R.~Usbeck, J.~Lehmann, and A.-C. Ngonga~Ngomo,
  ``Survey on challenges of question answering in the semantic web,''
  \emph{Semantic Web}, vol.~8, 11 2016.

\bibitem{sorokin2018modeling}
D.~Sorokin and I.~Gurevych, ``Modeling semantics with gated graph neural
  networks for knowledge base question answering,'' \emph{arXiv preprint
  arXiv:1808.04126}, 2018.

\bibitem{QAnswer}
D.~Diefenbach, A.~Both, K.~Singh, and P.~Maret, ``Towards a question answering
  system over the semantic web,'' \emph{Semantic Web}, vol.~11, pp. 421--439,
  2020.

\bibitem{MQALD}
L.~Siciliani, P.~Basile, P.~Lops, and G.~Semeraro, ``{MQALD}: Evaluating the
  impact of modifiers in question answering over knowledge graphs,''
  \emph{Semantic Web}, vol. Pre-press, pp. 1--17, 09 2021.

\bibitem{Wikidata}
D.~Vrande\v{c}i\'{c} and M.~Kr\"{o}tzsch, ``Wikidata: A free collaborative
  knowledgebase,'' \emph{Communications of the ACM}, vol.~57, no.~10, p.
  78–85, Sep. 2014.

\bibitem{keysers2019measuring}
D.~Keysers, N.~Sch{\"a}rli, N.~Scales, H.~Buisman, D.~Furrer, S.~Kashubin,
  N.~Momchev, D.~Sinopalnikov, L.~Stafiniak, T.~Tihon \emph{et~al.},
  ``Measuring compositional generalization: A comprehensive method on realistic
  data,'' \emph{arXiv preprint arXiv:1912.09713}, 2019.

\bibitem{bollacker2008freebase}
K.~Bollacker, C.~Evans, P.~Paritosh, T.~Sturge, and J.~Taylor, ``Freebase: a
  collaboratively created graph database for structuring human knowledge,'' in
  \emph{Proceedings of the 2008 ACM SIGMOD international conference on
  Management of data}, 2008, pp. 1247--1250.

\bibitem{Gerbil}
R.~Usbeck, M.~R\"{o}der, A.-C. Ngonga~Ngomo, C.~Baron, A.~Both, M.~Br\"{u}mmer,
  D.~Ceccarelli, M.~Cornolti, D.~Cherix, B.~Eickmann, P.~Ferragina, C.~Lemke,
  A.~Moro, R.~Navigli, F.~Piccinno, G.~Rizzo, H.~Sack, R.~Speck, R.~Troncy,
  J.~Waitelonis, and L.~Wesemann, ``Gerbil: General entity annotator
  benchmarking framework,'' in \emph{Proceedings of the 24th International
  Conference on World Wide Web}, ser. WWW '15.\hskip 1em plus 0.5em minus
  0.4em\relax Republic and Canton of Geneva, CHE: International World Wide Web
  Conferences Steering Committee, 2015, p. 1133–1143.

\bibitem{deeppavlov}
M.~Burtsev, A.~Seliverstov, R.~Airapetyan, M.~Arkhipov, D.~Baymurzina,
  N.~Bushkov, O.~Gureenkova, T.~Khakhulin, Y.~Kuratov, D.~Kuznetsov,
  A.~Litinsky, V.~Logacheva, A.~Lymar, V.~Malykh, M.~Petrov, V.~Polulyakh,
  L.~Pugachev, A.~Sorokin, M.~Vikhreva, and M.~Zaynutdinov, ``{D}eep{P}avlov:
  Open-source library for dialogue systems.''\hskip 1em plus 0.5em minus
  0.4em\relax Melbourne, Australia: Association for Computational Linguistics,
  2018, pp. 122--127.

\bibitem{tanon2018demoing}
T.~P. Tanon, M.~D. de~Assuncao, E.~Caron, and F.~M. Suchanek, ``Demoing
  {Platypus}--a multilingual question answering platform for {Wikidata},'' in
  \emph{European Semantic Web Conference}.\hskip 1em plus 0.5em minus
  0.4em\relax Springer, 2018, pp. 111--116.

\end{thebibliography}
%\begin{thebibliography}{1}


%\end{thebibliography}

\end{document}